\documentclass[letterpaper,journal]{IEEEtran}

\usepackage{amsmath,amsfonts,amssymb}
\usepackage{algorithm}
\usepackage{algpseudocode}
\usepackage{array}
\usepackage{textcomp}
\usepackage{stfloats}
\usepackage{url}
\usepackage{graphicx}
\usepackage{cite}
\usepackage{booktabs}

\newtheorem{remark}{Remark}
\newtheorem{proposition}{Proposition}

\begin{document}

\title{State-Augmented Learning for Post-Deployment Adaptation of Coexisting O-RAN Applications}

\author{Wonhyeok~Ko,~\IEEEmembership{Graduate Student Member,~IEEE}, Yeonseo~Jeong,~\IEEEmembership{Graduate Student Member,~IEEE}, and Songnam~Hong,~\IEEEmembership{Senior Member,~IEEE}%
\thanks{W. Ko and Y. Jeong contributed equally to this work.}
\thanks{W. Ko and S. Hong are with the Department of AI-Semiconductor, Hanyang University, Seoul, Korea (e-mail: blackhawk77@hanyang.ac.kr). Y. Jeong and S. Hong are with the Department of Electronic Engineering, Hanyang University, Seoul, Korea (e-mail: wjddustj225@hanyang.ac.kr; snhong@hanyang.ac.kr). Corresponding author: Songnam Hong.}}

\markboth{IEEE Transactions on Mobile Computing}%
{Ko \MakeLowercase{\textit{et al.}}: State-Augmented Learning for Post-Deployment Adaptation of Coexisting O-RAN Applications}

\maketitle

\begin{abstract}
Fixed control policies in Open RAN can require retraining when service requirements, traffic loads, or interacting applications change. We propose ATTUNE, a state-augmented learning method for an energy saving application that coexists with a learned mobility load balancing application under a service constraint. ATTUNE conditions the energy saving policy on a Lagrange multiplier and updates this scalar during operation using measured service degradation, while keeping both applications' network parameters fixed. To train a single policy across service requirements, ATTUNE samples target service levels uniformly and maps them to multipliers through an adaptive table. This procedure accounts for policy-dependent multiplier responses and improves energy savings compared with training over a widened fixed multiplier range. We characterize the induced distribution of service levels and the convergence and fluctuations of the online update. An exact finite-window identity establishes convergence of average measured degradation to the target when the projection corrections vanish on average. Simulations show that a single ATTUNE policy achieves energy savings comparable to policies trained separately for each service requirement. It adapts to changes in the target, load, and partner application without retraining, increases energy savings at light load, and maintains target tracking for separately trained application pairs. No statistically significant increase in cell switching is detected in the evaluated comparisons.
\end{abstract}

\begin{IEEEkeywords}
O-RAN, energy saving, load balancing, constrained reinforcement learning, state augmentation, multi-agent reinforcement learning.
\end{IEEEkeywords}

\section{Introduction}
\label{sec:introduction}

\IEEEPARstart{E}{nergy} saving through cell deactivation and mobility load balancing are key control functions implemented by Open RAN (O-RAN) applications on RAN Intelligent Controllers (RICs)~\cite{polese2023understanding,oran_usecases}. An energy saving rApp (ES-rApp) on the Non-Real-Time RIC controls cell activation, while a mobility load balancing xApp (MLB-xApp) on the Near-Real-Time RIC adjusts cell individual offsets (CIOs) to redistribute traffic. These applications may originate from different vendors and operate at different time scales~\cite{akman2024energy,hoffmann2024lessons}. Their actions are coupled through cell loads: deactivation redistributes traffic, while CIO control changes the load and resource demand of the remaining active cells. Our previous work~\cite{jeong2026accord} established sufficient conditions for oscillations under myopic ES and MLB rules despite the existence of a feasible operating point. Their interaction was formulated as a constrained Markov game with \emph{asymmetric} service coupling: MLB penalizes service degradation in its reward, whereas ES minimizes energy consumption subject to a service constraint. The resulting ACCORD algorithm trains the applications with separate rewards and critics, applying a Lagrange multiplier only to ES and requiring no exchange of control parameters.

ACCORD deploys fixed policies and discards its training multiplier and critics. Consequently, each policy is calibrated to a prescribed service requirement and training environment. This creates three deployment limitations. First, different service requirements require separately trained policies. Second, under light traffic, service degradation can remain substantially below the permitted level, limiting achievable energy savings; under overload, a separate guard is needed. Third, replacing the MLB partner can violate the service constraint, requiring ES recalibration and validation before deployment. These limitations motivate a policy that can adapt its operating point after training.

We propose ATTUNE for post-deployment adaptation of coexisting ES and MLB applications. ATTUNE preserves the asymmetric service coupling of ACCORD and conditions the ES policy on the service-constraint multiplier. During operation, the multiplier is updated using measured service degradation while the network parameters of both applications remain fixed. During training, service targets are sampled uniformly and mapped to multipliers through an adaptive table. Building on state-augmented constrained reinforcement learning~\cite{calvofullana2024state,naderializadeh2022state}, ATTUNE addresses three design challenges: selecting training multipliers that cover the desired service levels despite a policy-dependent response; training the MLB partner under the activation patterns induced by these multipliers, without providing it with the multiplier input; and adapting the operating point while controlling cell switching under dwell-time constraints.

\subsection{Related Work}

Primal-dual methods jointly update a policy and a Lagrange multiplier to enforce constraints~\cite{tessler2019rcpo,ray2019benchmarking}. Zero-duality-gap results establish the validity of dual optimization under the stated regularity conditions, with approximation guarantees for sufficiently expressive policy parametrizations~\cite{paternain2019zero}. ACCORD applies primal-dual learning to the ES agent within an asymmetrically coupled game and discards the multiplier after training~\cite{jeong2026accord}.

State augmentation incorporates dual variables into the policy input to enable adaptation during execution. Calvo-Fullana \emph{et al.}~\cite{calvofullana2024state} establish limitations of fixed reward scalarization in constrained problems and develop multiplier-conditioned policies with online dual dynamics. NaderiAlizadeh \emph{et al.}~\cite{naderializadeh2022state} apply state augmentation to radio resource management with long-term per-user constraints. Their policies receive network states and dual variables, which are sampled from a fixed distribution during training and updated periodically during execution. Agorio \emph{et al.}~\cite{agorio2024multi} extend state augmentation to multi-agent assignment, using dual-variable dynamics to coordinate task allocation. These formulations do not address the combination of asymmetric ES--MLB coupling, cell switching costs, and minimum dwell times considered here.

Yao \emph{et al.}~\cite{yao2023ccpo} propose constraint-conditioned policy optimization (CCPO), which conditions a policy on the constraint threshold to support multiple safety requirements. Training uses data from selected thresholds, threshold-dependent value estimates, and associated multipliers. At deployment, the policy receives the prescribed threshold without measurement-based input correction. CCPO considers an episodic discounted-cost constraint, whereas our setting imposes a long-run average service constraint on coexisting ES and MLB applications.

Table~\ref{tab:related} summarizes differences in policy inputs, training conditions, and deployment adaptation. ATTUNE combines adaptive target-uniform multiplier sampling with ES-only state augmentation, while retaining the MLB agent's separate reward and observation. This design addresses requirements arising from ES--MLB coexistence: the multiplier-to-service response varies across trained policies, and the multipliers sampled for ES affect the activation patterns encountered by the MLB agent, which does not observe the multiplier. In addition, online adaptation changes discrete activation decisions subject to switching costs and minimum dwell times. These coupled training and operational requirements are not jointly addressed by the related methods.

\begin{table}[t]
\centering
\caption{Comparison of policy inputs, training conditions, and deployment updates of related methods and ATTUNE.}
\label{tab:related}
\footnotesize
\setlength{\tabcolsep}{2.5pt}
\renewcommand{\arraystretch}{1.1}
\begin{tabular}{@{}
>{\raggedright\arraybackslash}p{0.23\columnwidth}
>{\raggedright\arraybackslash}p{0.19\columnwidth}
>{\raggedright\arraybackslash}p{0.28\columnwidth}
>{\raggedright\arraybackslash}p{0.23\columnwidth}@{}}
\toprule
Method & Policy input & Training conditions & Deployment update \\
\midrule
ACCORD~\cite{jeong2026accord}
& Observation
& Multiplier updated for fixed $V_{\max}$
& None; fixed policy \\
\midrule[0.3pt]

Calvo-Fullana \emph{et al.}~\cite{calvofullana2024state}
& State, multipliers
& Sampled multipliers; distribution not prescribed
& Online dual update \\
\midrule[0.3pt]

NaderiAlizadeh \emph{et al.}~\cite{naderializadeh2022state}
& State, dual variables
& Fixed multiplier sampling distribution
& Periodic dual update \\
\midrule[0.3pt]

CCPO~\cite{yao2023ccpo}
& State, threshold
& Selected training thresholds
& None; prescribed threshold input \\
\midrule[0.3pt]

ATTUNE
& Observation, multiplier (ES only)
& Uniform service targets mapped to multipliers via an adaptive table
& Measurement-based multiplier update; fixed network parameters \\
\bottomrule
\end{tabular}
\end{table}

\subsection{Contributions}

Our contributions are as follows.

\begin{itemize}
\item We develop a state-augmented framework for post-deployment adaptation of asymmetrically coupled O-RAN applications. ATTUNE conditions only the ES policy on the service-constraint multiplier and updates this scalar from measured degradation during operation. The MLB agent retains its own reward and observation, and both applications operate without network-parameter updates or control-parameter exchange.

\item We propose target-uniform multiplier sampling using an adaptive target-to-multiplier table. The table accounts for policy-dependent service responses and allocates training samples across the desired service range. We characterize the service-level distribution induced by log-uniform sampling and experimentally show that the proposed distribution improves ES performance and the effectiveness of the learned MLB partner relative to a widened fixed range.

\item We analyze the operational multiplier update to characterize target tracking and guide the selection of the update step size and period. Under the stated response and noise assumptions, diminishing-step updates converge to the target operating point or a range boundary. For constant-step operation, a local analysis quantifies the trade-off between response time and multiplier fluctuations, with a small-step limit connecting these results to the nonlinear projected update. An exact finite-window identity expresses average tracking error through the net multiplier change and projection corrections, while a dwell-time bound limits switching independently of the multiplier trajectory.

\item We evaluate adaptation across service requirements, traffic loads, and separately trained partners, including sequential changes during operation. A single ATTUNE policy achieves energy savings comparable to three specialized policies and approximately seven percentage points higher savings at light load than ACCORD. All 20 cross-seed pairs trained under the same load track their targets within $\pm0.005$ without retraining, and a target margin yields service-feasible operation for all these pairs. Comparisons with fixed and feedback-corrected policy inputs isolate the role of measurement feedback, and no statistically significant increase in switching is detected in the evaluated online-update comparisons.
\end{itemize}

Section~\ref{sec:system} summarizes the network model and the constrained Markov game. Section~\ref{sec:formulation} formulates the state-augmented problem, Section~\ref{sec:algorithm} presents the training and operation procedures, and Section~\ref{sec:analysis} analyzes them. Section~\ref{sec:evaluation} evaluates the approach, and Section~\ref{sec:conclusion} concludes the paper.

\section{System Model and Coexistence Game}
\label{sec:system}

This section summarizes the network model and asymmetrically constrained Markov game developed in~\cite{jeong2026accord}. The game provides the mathematical foundation for ATTUNE.

\subsection{Network Model and Performance Metrics}
A cluster of $B$ cells serves $U$ user equipment (UE) devices in slots of duration $T_0$. Traffic follows a diurnal profile that repeats every cycle of $T$ slots, compressed relative to a physical day, and one episode spans one cycle. The ES-rApp sets the activation vector $\mathbf{s}_t\in\{0,1\}^B$ once every $K$ slots, subject to a minimum dwell time $T_{\mathrm{dwell}}>K$ between state transitions of a cell and to the condition that every sleeping cell retains an active neighbor. The MLB-xApp adjusts the CIO vector $\boldsymbol{\theta}_t$ every slot by at most one quantization step per cell. Each UE in coverage associates with the active cell that maximizes its CIO-biased received power. A UE $u$ with traffic demand $d_u$ and achievable rate $r_{u,t}$ per physical resource block (PRB) requires $d_u/r_{u,t}$ PRBs to satisfy its demand. The normalized load of cell $b$, with $N_b$ available PRBs and associated UE set $\mathcal{U}_{b,t}$, is
\begin{equation}
  \rho_{b,t}=\frac{1}{N_b}\sum_{u\in\mathcal{U}_{b,t}}\frac{d_u}{r_{u,t}}.
  \label{eq:load}
\end{equation}
Each UE $u$ served by cell $b$ receives a rate of $d_u/\max\{\rho_{b,t},1\}$, with its PRB demand proportionally reduced when the cell is overloaded.

We adopt an affine load-dependent power consumption model with an additional cell-switching cost. The network power consumption under activation vector $\mathbf{s}_t$ is
\begin{equation}
  \begin{aligned}
  E_t={}&\sum_b\Bigl[s_{b,t}\bigl(P^{\mathrm{fix}}_b
    +\Delta_b\min\{\rho_{b,t},1\}P^{\mathrm{tx}}_b\bigr)\\
    &\qquad +(1-s_{b,t})P^{\mathrm{sleep}}_b\Bigr]
    +\zeta\sum_b|s_{b,t}-s_{b,t-1}|.
  \end{aligned}
  \label{eq:power}
\end{equation}
The terms account for load-independent active power $P^{\mathrm{fix}}_b$, load-dependent power with slope $\Delta_b$ relative to the transmit power $P^{\mathrm{tx}}_b$, sleep-state power $P^{\mathrm{sleep}}_b$, and switching power. The coefficient $\zeta$ is the transition energy divided by $T_0$, and $\bar E_t$ denotes the network power consumption normalized by the all-active, fully loaded power without transitions. Service degradation is defined as the fraction of UEs that are in outage or receive less than a fraction $\beta\in(0,1]$ of their traffic demand:
\begin{equation}
  V_t=\frac{1}{U}\Bigl(|\mathcal{U}^{\mathrm{out}}_t|+\sum_{u\notin\mathcal{U}^{\mathrm{out}}_t}\mathbf{1}\bigl[\rho_{b^\star(u),t}>1/\beta\bigr]\Bigr),
  \label{eq:violation}
\end{equation}
where $\mathcal{U}^{\mathrm{out}}_t$ denotes the set of UEs not covered by any active cell and $b^\star(u)$ denotes the serving cell of UE $u$. The handover rate $H_t$ is the fraction of UEs that change their serving cell in slot $t$. 

\subsection{Asymmetrically Constrained Markov Game}
\label{subsec:game}

Proposition~1 of~\cite{jeong2026accord} identifies a two-cell configuration in which load balancing reduces a cell's load below the ES deactivation threshold, while subsequent deactivation overloads the remaining cell. The myopic rules then alternate between activation states despite a service-feasible balanced configuration. Proposition~2 of~\cite{jeong2026accord} further shows that identical pre-deactivation observations can lead to different post-deactivation service outcomes. These findings motivate accounting for the service consequences of the applications' coupled actions. Following~\cite{jeong2026accord}, we model ES--MLB coexistence as an asymmetrically constrained Markov game with agents $\mathrm{E}$ and $\mathrm{M}$ representing ES and MLB, respectively. ES acts at the start of epoch $k$ using observation $\mathbf{o}^{\mathrm{E}}_k$, whereas MLB acts in each slot $t$ using observation $\mathbf{o}^{\mathrm{M}}_t$. Both observations use preceding-slot measurements. ES observes cell loads, demand histories, its activation vector, and dwell timers; MLB observes cell loads, demand histories, and its CIO vector. Neither agent observes the other application's control configuration.

The MLB agent receives the slot-level reward $R^{\mathrm{M}}_t=-V_t-w_{\mathrm{h}}H_t$, where $w_{\mathrm{h}}$ weights the handover penalty. The ES agent receives the epoch-level reward $R^{\mathrm{E}}_k=-\frac{1}{K}\sum_{t\in\mathcal{T}_k}\bar E_t$, where $\mathcal{T}_k=\{kK,\dots,(k+1)K-1\}$. The operator imposes the long-run average service constraint
\begin{equation}
  C(\pi)=\lim_{T'\to\infty}\frac{1}{T'}\sum_{t=0}^{T'-1}\mathbb{E}^{\pi}[V_t]\le V_{\max}
  \label{eq:constraint}
\end{equation}
on the joint policy $\pi=(\pi^{\mathrm{E}},\pi^{\mathrm{M}})$. The coupling is \emph{asymmetric}: MLB directly penalizes service degradation in its reward, whereas ES minimizes energy consumption subject to the service constraint. The desired interaction is characterized by a generalized Nash equilibrium: a service-feasible joint policy at which ES cannot improve its return through a service-feasible unilateral deviation, while MLB cannot improve its return through any admissible unilateral deviation~\cite{jeong2026accord,facchinei2010gne}.

To handle the service constraint in this game, ACCORD introduces a Lagrange multiplier $\lambda\ge0$ into the ES policy update and uses the penalized advantage
\begin{equation}
\begin{aligned}
  \hat A^{\mathrm{E}}_k
  =\frac{1}{1+\lambda}\left[
  \frac{A^{\mathrm{E}}_k-\mu_{\mathrm{E}}}{\sigma_{\mathrm{E}}}
  -\lambda\frac{A^C_k-\mu_C}{\sigma_C}
  \right].
\end{aligned}
\label{eq:penalized_adv}
\end{equation}
Here, $A^{\mathrm{E}}_k$ and $A^C_k$ denote the energy-reward and service-cost advantages, $\mu_{\mathrm{E}}$ and $\mu_C$ their respective means, and $\sigma_{\mathrm{E}}$ and $\sigma_C$ their respective standard deviations used for normalization.

\begin{figure*}[t]
\centering
\includegraphics[width=\textwidth]{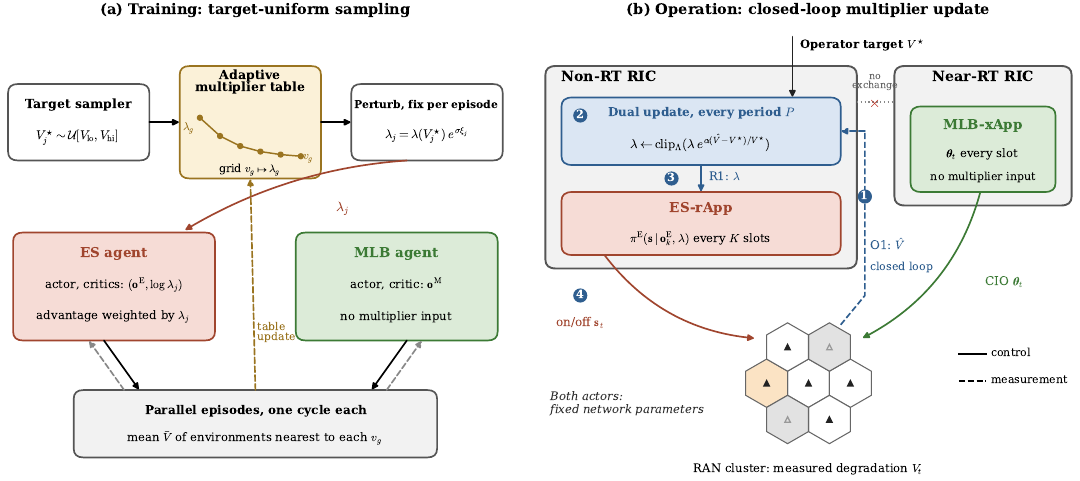}
\caption{Overview of ATTUNE. (a)~Training: each environment uniformly samples a service target, maps it to a multiplier through an adaptive table, and perturbs the multiplier. The multiplier is held fixed throughout the episode, and its logarithm conditions the ES actor and critics; the MLB agent receives no multiplier input. Measured degradation updates the table. (b)~Closed-loop operation: (1)~measure service degradation, (2)~update the multiplier from its deviation from the operator's target, (3)~supply the multiplier to the ES-rApp, and (4)~apply activation decisions that affect subsequent degradation. Both actors retain fixed network parameters, and the applications exchange no control configurations.}
\label{fig:system}
\end{figure*}

\section{State-Augmented Formulation}
\label{sec:formulation}

ATTUNE extends the coexistence game of Section~\ref{subsec:game} by conditioning the ES policy on the service-constraint multiplier. In ACCORD, the multiplier is used only during training; in ATTUNE, it also serves as a policy input that is updated during operation while both agents' network parameters remain fixed, as illustrated in Fig.~\ref{fig:system}. This enables a single trained policy pair to adapt to different service targets and changing operating conditions without retraining.

\subsection{Augmented Energy Saving Policy}
\label{subsec:augmented}

The agents use the rewards defined in Section~\ref{subsec:game}. We augment the ES observation with the multiplier, yielding the policy $\pi^{\mathrm{E}}(\mathbf{s}\mid\mathbf{o}^{\mathrm{E}}_k,\lambda)$. For each $\lambda$, the ES agent aims to maximize the Lagrangian return given the MLB policy:
\begin{equation}
  \mathcal{L}^{\mathrm{E}}_{\lambda}(\pi^{\mathrm{E}};\pi^{\mathrm{M}})
  =J^{\mathrm{E}}(\pi)-\lambda J^{C}(\pi),
  \label{eq:lagrangian}
\end{equation}
where $\pi=(\pi^{\mathrm{E}},\pi^{\mathrm{M}})$ and
\begin{equation}
  J^{\mathrm{E}}(\pi)=\mathbb{E}^{\pi}\Bigl[\sum_{k=0}^{\infty}\gamma_{\mathrm{E}}^{k}R^{\mathrm{E}}_k\Bigr],
  \quad
  J^{C}(\pi)=\mathbb{E}^{\pi}\Bigl[\sum_{t=0}^{\infty}\gamma^{t}V_t\Bigr]
  \label{eq:returns}
\end{equation}
are the discounted energy-reward and service-cost returns, accumulated over epochs and slots, respectively, with discount factors $\gamma_{\mathrm{E}},\gamma\in(0,1)$. The cost return is a discounted surrogate for the long-run average degradation in \eqref{eq:constraint}, and the Lagrangian return expresses the trade-off between energy saving and service degradation.

In ACCORD, the multiplier is a training-time optimization variable used to learn a policy for a fixed service requirement. ATTUNE additionally treats the multiplier as a conditioning variable, learning a family of ES control behaviors represented by a single policy network. Each sampled multiplier conditions the policy and weights the service-cost advantage in \eqref{eq:penalized_adv}, enabling the operating point to be adjusted after training by changing the input rather than the network parameters. The ES value and cost critics are also conditioned on $\lambda$ because the policy they evaluate depends on it.

Conditioning the policy on $\lambda$ allows its operating point to be adjusted, but does not by itself identify the multiplier needed to meet a given service target. The degradation attained at a given $\lambda$ depends on the trained policy and operating conditions. Moreover, separate advantage normalization in \eqref{eq:penalized_adv} makes the relative cost weight in the proximal policy optimization (PPO) update~\cite{schulman2017ppo} equal to $\lambda\sigma_{\mathrm E}/\sigma_C$. ATTUNE therefore uses measured degradation to determine suitable multiplier inputs. During training, the adaptive table in Fig.~\ref{fig:system}(a) learns a mapping from service targets to multipliers. Targets are sampled uniformly, and each mapped multiplier is perturbed and held fixed throughout the episode; the resulting degradation measurements update the table. During operation, the feedback loop in Fig.~\ref{fig:system}(b) corrects the multiplier input according to the deviation of measured degradation from the prescribed target, while both actors' network parameters remain fixed.

The MLB agent maximizes $J^{\mathrm{M}}(\pi)=\mathbb{E}^{\pi}[\sum_{t=0}^{\infty}\gamma^tR^{\mathrm{M}}_t]$ using its own observation, without access to $\lambda$. Both agents are trained concurrently with separate rewards, while multiplier conditioning and multiplier-based penalization are applied only to ES.

\subsection{Online Multiplier Update}
\label{subsec:update}

We specify the measurement-based multiplier update used to adjust the operating point of the trained policy pair during deployment. For fixed operating conditions and an MLB policy $\pi^{\mathrm M}$, define $\bar V(\lambda)=C\bigl(\pi^{\mathrm E}_{\lambda},\pi^{\mathrm M}\bigr)$, where $\pi^{\mathrm E}_{\lambda}=\pi^{\mathrm E}(\cdot\mid\cdot,\lambda)$ denotes the trained ES policy with its multiplier input held fixed at $\lambda$. Since the response $\bar V(\cdot)$ depends on the trained ES policy, traffic load, and MLB partner, a multiplier calibrated under the training conditions may be unsuitable after deployment. Let $V^{\star}\in(0,V_{\max}]$ denote the service-degradation target specified by the operator. Assuming that $\bar V(\lambda)$ decreases with $\lambda$, we seek a multiplier $\lambda^{\star}\in\Lambda$ satisfying $\bar V(\lambda^{\star})=V^{\star}$, provided such a multiplier exists.

To track this target without explicitly estimating $\bar V(\cdot)$, ATTUNE updates the multiplier using the measured deviation from the target:
\begin{equation}
  \lambda_{n+1}=\mathrm{clip}_{\Lambda}\Bigl(
  \lambda_n\exp\bigl(\alpha\,(\hat V_n-V^{\star})/V^{\star}\bigr)
  \Bigr),
  \label{eq:dual_operation}
\end{equation}
where $n$ indexes update periods, $\hat V_n$ is the average degradation measured over the $n$th period, and $\alpha>0$ is the step size. The deviation is normalized by the target $V^{\star}$. The operator $\mathrm{clip}_{\Lambda}$ projects onto the operational range $\Lambda=[\lambda_{\min},\lambda_{\max}]$. Before projection, the update increases $\lambda$ for positive deviations, decreases it for negative deviations, and leaves it unchanged for zero deviation. The multiplicative update implements integral control in $\log\lambda$, accommodating multiplier values that span more than an order of magnitude across service levels.

The operational range $\Lambda=[\lambda_{\min},\lambda_{\max}]$ is determined from the final training table $\{\lambda_g\}$ as
\begin{equation}
\begin{aligned}
  \lambda_{\min}
  &=\max\left\{\underline\lambda,\ e^{-2\sigma}\min_g\lambda_g\right\},\\
  \lambda_{\max}
  &=\min\left\{\bar\lambda,\ e^{2\sigma}\max_g\lambda_g\right\},
\end{aligned}
\label{eq:operational_range}
\end{equation}
where $\lambda_g$ is the multiplier at grid point $g$ in the final training table, $\underline\lambda$ and $\bar\lambda$ are the training bounds, and $\sigma$ is the prescribed standard deviation of the log-space multiplier perturbation used during training. The range extends the table's log-multiplier range by $2\sigma$ at each end while respecting these bounds; Section~\ref{subsec:training} describes the table and perturbation. The update can start from any $\lambda_0\in\Lambda$, with the table value corresponding to $V^{\star}$ providing an initial estimate under the training conditions. Section~\ref{sec:evaluation} specifies the initialization used in each experiment.

\subsection{Design Requirements}\label{subsec:design}

State-augmented learning samples multipliers during training and holds each multiplier fixed within an episode~\cite{calvofullana2024state,naderializadeh2022state}. Applying this framework to ES--MLB coexistence requires addressing the following issues in training and operation.

The training distribution of multipliers should provide adequate coverage of the service targets of interest. Because $\bar V(\lambda)$ varies across trained policies, a fixed multiplier range can provide insufficient target coverage or allocate many samples outside the desired service range.

Multiplier sampling also affects the learning of the MLB partner. Only ES observes the multiplier, but both agents experience the resulting activation patterns. Excessive sampling outside the desired service range can therefore cause a mismatch between the activation patterns encountered by MLB during training and deployment.

The operational update must account for cell switching behavior. Measurement variability can cause the multiplier to fluctuate around its target operating point, changing activation decisions and inducing reversals even when the dwell constraint is satisfied.

\section{ATTUNE: Training and Operation}
\label{sec:algorithm}

This section presents ATTUNE's training and operation procedures to address the design requirements in Section~\ref{subsec:design}. We describe how uniformly sampled service targets are mapped to training multipliers, how the online update parameters are selected, and how overload is handled.

\subsection{Training}
\label{subsec:training}

ATTUNE jointly trains the ES and MLB agents using PPO, with an adaptive table that maps service targets to ES multipliers. After warm-up, each training iteration consists of target sampling and multiplier assignment, rollout collection and PPO updates, and a table update based on measured service degradation. The table $\{(v_g,\lambda_g)\}_{g=1}^{G}$ is defined on a target grid $V_{\mathrm{lo}}=v_1<\dots<v_G=V_{\mathrm{hi}}$, where $[V_{\mathrm{lo}},V_{\mathrm{hi}}]$ covers the service levels of interest with a margin on each side.

We initialize the target-to-multiplier table using service-degradation measurements collected during warm-up. In this phase, multipliers are sampled log-uniformly from a fixed range and used to train the agents. Using measurements from the final warm-up iterations, we fit degradation as a linear function of $\log\lambda$ by least squares:
\begin{equation}
  \widehat V_{\mathrm{fit}}(\lambda)=q_0+q_1\log\lambda,
  \label{eq:warmup_fit}
\end{equation}
where $q_0$ and $q_1$ are the fitted intercept and slope. If $q_1<0$, the fit indicates that degradation decreases with the multiplier. We invert this fitted response to initialize the multiplier for each grid target $v_g$:
\begin{equation}
  \lambda_g=\mathrm{clip}_{[\underline\lambda,\bar\lambda]}
  \left(\exp\left(\frac{v_g-q_0}{q_1}\right)\right).
  \label{eq:table_init}
\end{equation}
If $q_1\ge0$, we instead initialize the table multipliers with uniform spacing in log space over the warm-up range, assigning larger multipliers to stricter targets and clipping all entries to $[\underline\lambda,\bar\lambda]$. Both cases provide the initial table for the subsequent adaptive updates. Section~\ref{subsec:setup} specifies the warm-up duration, sampling range, and fitting window.

After initialization, we use the current table to assign a training multiplier to each parallel environment. At each training iteration, environment $j$ samples a target $V_j^{\star}$ uniformly from $[V_{\mathrm{lo}},V_{\mathrm{hi}}]$. The corresponding multiplier is obtained by linear interpolation of the table entries in log space: for $v\in[v_g,v_{g+1}]$,
\begin{equation}
  \begin{aligned}
  \log\lambda(v)
  ={}&\frac{v_{g+1}-v}{v_{g+1}-v_g}\log\lambda_g +\frac{v-v_g}{v_{g+1}-v_g}\log\lambda_{g+1}.
  \end{aligned}
  \label{eq:lambda_interp}
\end{equation}
The interpolated multiplier is perturbed multiplicatively,
\begin{equation}
  \lambda_j=\lambda(V_j^{\star})\exp(\sigma\xi_j),
  \qquad \xi_j\sim\mathcal{N}(0,1),
  \label{eq:lambda_sample}
\end{equation}
and then clipped to $[\underline\lambda,\bar\lambda]$. The perturbation provides training coverage around the multiplier associated with each target. We use $\sigma=0.5$, chosen before training ATTUNE from the standard deviation of $\log\lambda$ measured during online operation of the policy trained log-uniformly over the warm-up range, with $\alpha=0.5$ and $V^\star=0.05$ ($0.52$); the trained ATTUNE policy shows $0.41$ under the same update. The resulting $\lambda_j$ is held fixed throughout the environment's episode.

We use the assigned multipliers to train the ES policy and critics jointly with the MLB agent. The ES actor and both ES critics receive $\log\lambda_j$ as an additional input. An affine transformation maps the logarithms of the warm-up range endpoints to $0$ and $1$, without clipping the transformed input. For each environment $j$, the ES actor update uses the penalized advantage in \eqref{eq:penalized_adv} with the same $\lambda_j$ used to collect its rollout. The ES value and cost critics learn the corresponding reward and service-cost returns. MLB is trained concurrently using its own observations and reward, without multiplier conditioning. The base network architectures and PPO update procedure follow ACCORD~\cite{jeong2026accord}, with the additional ES inputs described above.

We then update the table using measured degradation to improve the correspondence between each target and its assigned multiplier. After each training iteration, environments are grouped by the grid point nearest to their sampled targets. For each grid point $g$ with assigned environments, their mean degradation $\tilde V_g$ updates the smoothed estimate $\hat C_g\leftarrow(1-\kappa)\hat C_g+\kappa\tilde V_g$, where $\kappa$ is the smoothing coefficient. The table multiplier is adjusted according to the relative deviation from the grid target:
\begin{equation}
  \lambda_g\leftarrow\Pi_{[\underline\lambda,\bar\lambda]}
  \Bigl(\lambda_g\exp\Bigl[
  \mathrm{clip}_{[-\delta,\delta]}
  \Bigl(\eta\,\frac{\hat C_g-v_g}{v_g}\Bigr)
  \Bigr]\Bigr),
  \label{eq:table_update}
\end{equation}
where $\eta$ is the step size, $\delta$ limits the log-multiplier adjustment per iteration, and $\Pi_{[\underline\lambda,\bar\lambda]}$ projects onto the table bounds. Grid points without assigned environments skip this measurement-based update. After all grid-point updates, monotonicity is restored by a running minimum from the strictest target, $\lambda_g\leftarrow\min_{g'\le g}\lambda_{g'}$, allowing ties without imposing minimum spacing. The table thus adapts to the service response of the policies being trained, subject to the perturbation, grid resolution, and multiplier bounds. Algorithm~\ref{alg:training} summarizes the training procedure.

\begin{algorithm}[t]
\caption{ATTUNE Training}
\label{alg:training}
\begin{algorithmic}[1]
\Require Target range $[V_{\mathrm{lo}},V_{\mathrm{hi}}]$, grid $\{v_g\}$, noise scale $\sigma$, smoothing coefficient $\kappa$, table step size $\eta$, update limit $\delta$, and bounds $[\underline\lambda,\bar\lambda]$
\State Initialize both agents' actors and critics, with $\log\lambda$ as an additional input to the ES actor and critics
\State Initialize the multiplier table $\{\lambda_g\}$ through warm-up
\For{each post-warm-up training iteration}
  \For{each environment $j$}
    \State Draw $V^{\star}_j\sim\mathrm{Unif}[V_{\mathrm{lo}},V_{\mathrm{hi}}]$ and sample $\lambda_j$ using \eqref{eq:lambda_sample}; clip it to $[\underline\lambda,\bar\lambda]$ and hold it fixed throughout the episode
  \EndFor
  \State Collect rollouts in all environments using $\pi^{\mathrm{E}}(\cdot\mid\mathbf{o}^{\mathrm{E}},\lambda_j)$ and $\pi^{\mathrm{M}}(\cdot\mid\mathbf{o}^{\mathrm{M}})$
  \State Compute $\hat A^{\mathrm{E}}$ using \eqref{eq:penalized_adv} with each environment's $\lambda_j$; update both agents using PPO
  \State Update $\hat C_g$ and $\lambda_g$ using \eqref{eq:table_update} for grid points with assigned environments
  \State Restore monotonicity: $\lambda_g\leftarrow\min_{g'\le g}\lambda_{g'}$
\EndFor
\end{algorithmic}
\end{algorithm}

\begin{algorithm}[t]
\caption{ATTUNE Operation}
\label{alg:operation}
\begin{algorithmic}[1]
\Require Trained actors; requirement $V_{\max}$, margin $m$, step $\alpha$, period $P$, range $\Lambda$, initial $\lambda_0$
\State $V^{\star}\gets V_{\max}-m$; $\lambda\gets\mathrm{clip}_{\Lambda}(\lambda_0)$
\For{each slot $t$}
  \State ES-rApp: at the start of each epoch $k$, select the activation vector from $\pi^{\mathrm{E}}(\cdot\mid\mathbf{o}^{\mathrm{E}}_k,\lambda)$ with dwell mask and projection
  \State MLB-xApp: select $\mathbf{a}^{\mathrm{M}}_t$ from $\pi^{\mathrm{M}}(\cdot\mid\mathbf{o}^{\mathrm{M}}_t)$
  \State Measure $V_t$
  \If{$t$ ends an update period of length $P$}
    \State Update $\lambda$ by \eqref{eq:dual_operation} with $\hat V$ averaged over the period
  \EndIf
\EndFor
\end{algorithmic}
\end{algorithm}

\subsection{Operation}
\label{subsec:operation}

During operation, both actors retain their trained network parameters. Only the scalar multiplier is updated using \eqref{eq:dual_operation}. The Non-RT RIC maintains the multiplier and passes it to the ES-rApp through the R1 interface. The MLB-xApp retains its original input and policy. Algorithm~\ref{alg:operation} summarizes the procedure. The following guidelines specify the update step size and period, the interpretation of range saturation, and the target margin used to accommodate finite-horizon variability.

\noindent\textit{Update parameters:} The step size $\alpha$ and update period $P$ jointly determine response time and multiplier variability. Let $u=\log\lambda$, $F(u)=\bar V(e^u)$, and $u^{\dagger}$ denote an interior target operating point. Near this point, Proposition~\ref{prop:fluct} gives an approximate time constant $\tau\approx P/(\alpha a)$, where $a=-F'(u^{\dagger})/V^{\star}>0$ is the local response sensitivity. Adjacent entries of the trained table with distinct multipliers provide a finite-difference estimate of $a$. For a desired $\tau$ shorter than the time scale of operating-condition changes, select $\alpha=P/(a\tau)$ subject to $\alpha a<2$. If the averaging period is long compared with the correlation time of $V_t$, the measurement-noise variance satisfies $\omega^2\approx\nu^2/P$ for a constant $\nu^2$. Substitution into \eqref{eq:var} gives $\mathrm{Var}(u)\approx\nu^2/[a^2(V^{\star})^2\tau(2-P/\tau)]$. Thus, $P$ should allow sufficient averaging while remaining short relative to $\tau$. The predicted standard deviation should also be compared with the distance from $u^{\dagger}$ to the nearer boundary of $\mathcal{U}=[\log\lambda_{\min},\log\lambda_{\max}]$, particularly at stringent targets. 

These local relations do not fully describe large changes in operating conditions. Before projection, each update changes $u$ by $\alpha(\hat V_n-V^{\star})/V^{\star}$, which is bounded below by $-\alpha$. A decrease of $\Delta$ in $u$ therefore requires at least $\lceil\Delta/\alpha\rceil$ updates, whereas an increase greater than $\alpha$ in one update requires $\hat V_n>2V^{\star}$. Section~\ref{subsec:setup} specifies the selected parameters, and Sections~\ref{subsec:res_closed} and~\ref{subsec:res_dynamic} assess the resulting response.

\noindent\textit{Operational range:} The multiplier is projected onto the range $\Lambda$ defined in Section~\ref{subsec:update}. Persistent boundary saturation together with a sustained deviation from the service target indicates a possible mismatch between the operating conditions and the response range of the trained policy.

\noindent\textit{Target margin:} The operator sets $V^{\star}=V_{\max}-m$, with $0\le m<V_{\max}$, to allow for finite-horizon variability and projection effects. The margin is selected using degradation measurements over the intended evaluation horizon. When projection is inactive, Proposition~\ref{prop:average} expresses the average tracking error through an endpoint term that vanishes as the horizon increases. Deployment validation therefore checks the margin and initial adaptation transient for the intended operating conditions and policy pairing.

\begin{remark}[Operation under overload]
\label{rem:overload}
When traffic demand increases beyond the training conditions, the service target may become unattainable by the trained policy over $\Lambda$. Persistent saturation at $\lambda_{\max}$ with degradation above the target indicates a possible need for recalibration or retraining. A traffic-based guard can provide a faster response to sudden overload by suspending cell deactivation when total traffic exceeds a selected percentile of its training distribution~\cite{jeong2026accord}. Section~\ref{subsec:res_deploy} evaluates the guard and multiplier adaptation separately.
\end{remark}

\section{Analysis}
\label{sec:analysis}

This section establishes how multiplier sampling affects service-target coverage and how the online update regulates service degradation. We characterize target tracking and range saturation, quantify the local trade-off between response time and multiplier fluctuations, and derive an exact identity for finite-window average tracking errors. We also bound cell switching under the dwell-time constraint. These results support the sampling design and the selection of operational parameters.

We consider a fixed trained ES--MLB policy pair under fixed traffic conditions. Let $u=\log\lambda$, and let $F(u)$ denote the mean service degradation when the multiplier is held at $\lambda=e^u$ under the operating protocol considered. For the cycle-based evaluation used in this paper, each cycle starts from a prescribed initial state and spans one update period, giving
\begin{equation}
  F(u)=\mathbb{E}\left[\frac{1}{T}\sum_{t=0}^{T-1}V_t
  \,\middle|\,
  \begin{array}{l}
    \text{prescribed initialization},\\
    \lambda=e^u
  \end{array}\right].
  \label{eq:F_def}
\end{equation}
With independent randomness across cycles, this expectation also equals the long-run average under the same initialization protocol, so $F(u)=\bar V(e^u)$ with $\bar V$ defined in Section~\ref{subsec:update}. For continuous operation without reinitialization, $F(u)$ instead denotes the long-run average response; the expected degradation over an individual update period can also depend on its initial network state. Let $\mathcal{U}=[u_{\min},u_{\max}]=[\log\lambda_{\min},\log\lambda_{\max}]$. The response-based results use the following assumptions, as specified in each proposition.
\begin{itemize}
\item[(A1)] $F$ is continuously differentiable on $\mathcal U$ with $F'(u)<0$.
\item[(A2)] The degradation measured over update period $n$ satisfies $\hat V_n=F(u_n)+\varepsilon_n$, where $\mathbb{E}[\varepsilon_n\mid u_0,\{\hat V_i\}_{i=0}^{n-1}]=0$ and $\mathbb{E}[\varepsilon_n^2]\le\omega^2<\infty$.
\end{itemize}
Assumption (A1) specifies a differentiable, decreasing service response, providing the negative-feedback structure of the multiplier update and the response slope used in the sampling-density and local-dynamics analyses. Assumption (A2) models the period-average degradation as a conditionally unbiased response measurement with bounded variance. Under the cycle-based evaluation protocol, prescribed initialization and independent randomness make the conditional mean equal to $F(u_n)$, while $0\le V_t\le1$ ensures bounded variance. Without cycle resets, (A2) is an approximation whose accuracy depends on the effects of initial-state transients and temporal correlations over the update period. The finite-window tracking identity and dwell-time switching bound derived below do not require (A1)--(A2).

\subsection{Service-Target Coverage}
\label{subsec:an_distribution}

The following result characterizes the service-level distribution induced by log-uniform multiplier sampling and provides the rationale for ATTUNE's target-uniform sampling.

\begin{proposition}[Service-level distribution]
\label{prop:density}
Under (A1), let $u$ be drawn uniformly from $\mathcal{U}$ and let $V=F(u)$ denote the corresponding mean service degradation. Then $V$ has density
\begin{equation}
  p_V(v)=\frac{1}{(u_{\max}-u_{\min})\,\bigl|F'\bigl(F^{-1}(v)\bigr)\bigr|}
  \label{eq:density}
\end{equation}
on $[F(u_{\max}),F(u_{\min})]$. For any service interval $[v_a,v_b]$ within this range, the fraction of samples assigned to that interval is
\begin{equation}
  \Pr\{V\in[v_a,v_b]\}
  =\frac{F^{-1}(v_a)-F^{-1}(v_b)}{u_{\max}-u_{\min}}.
  \label{eq:fraction}
\end{equation}
\end{proposition}

\begin{IEEEproof}
Since $F$ is strictly decreasing, for $v\in[F(u_{\max}),F(u_{\min})]$,
\begin{equation*}
  \Pr\{V\le v\}
  =\Pr\{u\ge F^{-1}(v)\}
  =\frac{u_{\max}-F^{-1}(v)}{u_{\max}-u_{\min}}.
\end{equation*}
Differentiation using $(F^{-1})'(v)=1/F'(F^{-1}(v))$ gives \eqref{eq:density}. Taking the difference of the distribution function at $v_b$ and $v_a$ gives \eqref{eq:fraction}.
\end{IEEEproof}

Equation~\eqref{eq:density} shows that the density of sampled mean service levels is inversely proportional to the response slope magnitude. Thus, service intervals of the same small width receive more samples where the response slope has small magnitude and fewer where it has large magnitude. In the evaluated policies, flat regions occur near stringent service levels, where most cells remain active. Moreover, because the response differs across trained policies, the same fixed multiplier range can allocate different fractions of samples to the desired service interval, as quantified by \eqref{eq:fraction}. Merely widening the multiplier range therefore does not ensure balanced coverage.

ATTUNE addresses this imbalance by sampling $V^{\star}$ uniformly and adapting the table that maps each target to a multiplier. With an exact inverse mapping and without perturbation or clipping, $u=F^{-1}(V^{\star})$ gives $F(u)=V^{\star}$, yielding uniform coverage of mean service levels over the attainable target interval. In practice, table approximation, policy updates during training, perturbation, and clipping modify this ideal distribution. Away from the multiplier bounds, a small log-space perturbation with standard deviation $\sigma$ produces a local service-level standard deviation of approximately $\sigma|F'(F^{-1}(V^{\star}))|$. Thus, the table directs samples toward the service range of interest, while the perturbation provides coverage around the mapped operating points.

\subsection{Target Tracking and Range Saturation}
\label{subsec:an_update}

The following result establishes target tracking with diminishing steps when the prescribed service level is attainable within the operational range and characterizes boundary saturation otherwise. Writing \eqref{eq:dual_operation} in $u=\log\lambda$ and allowing a diminishing step size gives
\begin{equation}
  u_{n+1}=\Pi_{\mathcal{U}}\Bigl(
  u_n+\frac{\alpha_n}{V^{\star}}(\hat V_n-V^{\star})
  \Bigr),
  \label{eq:dual_log}
\end{equation}
where $\Pi_{\mathcal{U}}$ projects onto $\mathcal{U}$. Since the exponential function is increasing, this projection is equivalent to clipping the multiplier to $\Lambda$.

\begin{proposition}[Target tracking and range saturation]
\label{prop:dual}
Under (A1) and (A2), let $u_0\in\mathcal{U}$ and let the step sizes satisfy $\alpha_n>0$, $\sum_n\alpha_n=\infty$, and $\sum_n\alpha_n^2<\infty$. Then $u_n\to u^{\dagger}$ almost surely, where
\begin{equation}
  u^{\dagger}=
  \begin{cases}
    u_{\min}, & V^{\star}\ge F(u_{\min}),\\
    F^{-1}(V^{\star}), & F(u_{\max})\le V^{\star}<F(u_{\min}),\\
    u_{\max}, & V^{\star}<F(u_{\max}).
  \end{cases}
  \label{eq:limit}
\end{equation}
Consequently, $F(u_n)\to F(u^{\dagger})$ almost surely, and the limiting degradation equals the target whenever $V^{\star}\in[F(u_{\max}),F(u_{\min})]$.
\end{proposition}
\begin{IEEEproof}
Let $h(u)=(F(u)-V^{\star})/V^{\star}$ and $Y_n=(\hat V_n-V^{\star})/V^{\star}$. Then \eqref{eq:dual_log} becomes $u_{n+1}=\Pi_{\mathcal{U}}(u_n+\alpha_nY_n)$, a projected stochastic approximation of the form in~\cite[(5.1.3)]{kushner2003stochastic}. We verify the conditions of~\cite[Thm.~5.2.1]{kushner2003stochastic}. The step-size assumptions imply $\alpha_n\to0$ and satisfy the required step-size conditions. By (A2), $\mathbb{E}[Y_n\mid u_0,Y_i,i<n]=h(u_n)$, since $Y_i$ is an invertible affine transformation of $\hat V_i$. Thus, the conditional measurement bias is zero. By (A1), $h$ is continuous and bounded on the compact interval $\mathcal{U}$, and
\begin{equation*}
  \mathbb{E}[Y_n^2]
  \le 2\max_{u\in\mathcal{U}}h(u)^2+\frac{2\omega^2}{(V^{\star})^2}<\infty.
\end{equation*}
These properties establish the continuity, moment, and noise conditions of the theorem.

For the associated projected ordinary differential equation, an interior point is stationary when $h(u)=0$, the lower boundary is stationary when $h(u_{\min})\le0$, and the upper boundary is stationary when $h(u_{\max})\ge0$. Since $F$ is strictly decreasing, these conditions give the unique stationary point $u^{\dagger}$ in \eqref{eq:limit}. Moreover, $h(u)>0$ for every $u<u^{\dagger}$ in $\mathcal{U}$ and $h(u)<0$ for every $u>u^{\dagger}$ in $\mathcal{U}$, so the drift points toward $u^{\dagger}$. Define $f(u)=-\int_{u_{\min}}^{u}h(v)\,dv$. Then $f$ is continuously differentiable, $h=-f'$, and $f$ is constant on the singleton stationary set $\{u^{\dagger}\}$, satisfying~\cite[(A5.2.6)]{kushner2003stochastic}. The projected stochastic approximation theorem therefore yields $u_n\to u^{\dagger}$ almost surely. Continuity of $F$ gives $F(u_n)\to F(u^{\dagger})$, which equals $V^{\star}$ whenever the target lies in $[F(u_{\max}),F(u_{\min})]$.
\end{IEEEproof}

Proposition~\ref{prop:dual} characterizes target attainment within the response range of the trained policy pair. If $V^{\star}\in[F(u_{\max}),F(u_{\min})]$, the diminishing-step update converges to a multiplier that achieves the target. A target below this range leads to upper-bound saturation with degradation exceeding the target by $F(u_{\max})-V^{\star}$, whereas a target above this range leads to lower-bound saturation with degradation below the target. Under upper-bound saturation and with $V^{\star}=V_{\max}-m$, the limiting degradation exceeds the service requirement if and only if $F(u_{\max})-V^{\star}>m$. These outcomes describe the adjustment available through the multiplier for the fixed trained pair and operational range.

ATTUNE uses a constant step size to remain responsive to changing operating conditions. The next subsection analyzes the resulting response time and fluctuations around an interior target. Proposition~\ref{prop:average} subsequently characterizes finite-window average tracking, including the effect of projection.

\subsection{Response Time and Multiplier Fluctuations}
\label{subsec:an_fluct}

We analyze the trade-off between response time and multiplier fluctuations under a constant step size through two complementary results. Proposition~\ref{prop:fluct} derives explicit formulas for a local affine approximation without projection. Proposition~\ref{prop:small_step} then establishes the corresponding leading-order behavior of the nonlinear projected update as the step size tends to zero.

\begin{proposition}[Local response time and multiplier fluctuations]
\label{prop:fluct}
Let $u^{\dagger}=F^{-1}(V^{\star})$ lie in the interior of $\mathcal{U}$, and suppose $F$ is differentiable at $u^{\dagger}$ with $a=-F'(u^{\dagger})/V^{\star}>0$. Consider the unprojected update in \eqref{eq:dual_log} with constant step $\alpha>0$ and the local response model
\begin{equation*}
  \hat V_n=V^{\star}-V^{\star}a(u_n-u^{\dagger})+\varepsilon_n,
\end{equation*}
where $\mathbb{E}[\varepsilon_n\mid\mathcal{F}_n]=0$, $\mathbb{E}[\varepsilon_n^2]=\omega^2>0$, and $\mathcal{F}_n$ is generated by $u_0$ and the preceding measurements. If $\mathbb{E}[u_0^2]<\infty$, the deviation $e_n=u_n-u^{\dagger}$ satisfies
\begin{equation}
  e_{n+1}=(1-\alpha a)e_n+\frac{\alpha}{V^{\star}}\varepsilon_n.
  \label{eq:ar1}
\end{equation}
For $0<\alpha a<2$, its mean converges to zero as $\mathbb{E}[e_n]=(1-\alpha a)^n\mathbb{E}[e_0]$, and its variance converges to
\begin{equation}
  \lim_{n\to\infty}\mathrm{Var}(e_n)
  =\frac{\alpha\omega^2}{(V^{\star})^2(2a-\alpha a^2)}.
  \label{eq:var}
\end{equation}
For $0<\alpha a<2$ with $\alpha a\neq1$, the decay envelope of the mean has time constant $-1/\log|1-\alpha a|$ updates, approximately $1/(\alpha a)$ for small $\alpha a$. For $\alpha a=1$, the mean becomes zero after one update.
\end{proposition}
\begin{IEEEproof}
Substituting the local response model into the unprojected update gives \eqref{eq:ar1}. Let $\phi=1-\alpha a$. Since $e_n$ is $\mathcal{F}_n$-measurable and the noise has zero conditional mean, $\mathbb{E}[\varepsilon_n]=0$ and $\mathbb{E}[e_n\varepsilon_n]=0$. Therefore,
\begin{equation*}
  \begin{aligned}
    \mathbb{E}[e_{n+1}]&=\phi\,\mathbb{E}[e_n],\\
    \mathrm{Var}(e_{n+1})
    &=\phi^2\mathrm{Var}(e_n)
      +\frac{\alpha^2\omega^2}{(V^{\star})^2}.
  \end{aligned}
\end{equation*}
When $|\phi|<1$, equivalently $0<\alpha a<2$, the mean converges to zero and the variance converges to $\alpha^2\omega^2/[(V^{\star})^2(1-\phi^2)]$, yielding \eqref{eq:var}. For $0<|\phi|<1$, the mean envelope decays as $|\phi|^n=\exp(n\log|\phi|)$, giving the stated time constant. The approximation follows from $-\log(1-\alpha a)=\alpha a+O((\alpha a)^2)$ for small $\alpha a$. If $\phi=0$, the mean is zero for all $n\ge1$.
\end{IEEEproof}

\begin{proposition}[Small-step limit of the projected update]
\label{prop:small_step}
Consider \eqref{eq:dual_log} with constant step $\alpha>0$ and $u_0\in\mathcal{U}$. Suppose (A1)--(A2) hold with a fixed response $F$ and a variance bound $\omega^2$ independent of $\alpha$. Let $u^{\dagger}=F^{-1}(V^{\star})$ lie in the interior of $\mathcal{U}$. Assume that the measurement noise is uniformly bounded over $n$ and $\alpha$ and satisfies
\begin{equation*}
  \mathbb{E}[\varepsilon_n^2\mid\mathcal{F}_n]
  =\varsigma^2(u_n),
\end{equation*}
where $\mathcal{F}_n$ is generated by $u_0$ and the preceding measurements, and $\varsigma^2$ is continuous, independent of $\alpha$, and positive at $u^{\dagger}$. Let $a=-F'(u^{\dagger})/V^{\star}>0$, and choose deterministic integer indices $q_\alpha$ such that $\alpha q_\alpha/|\log\alpha|\to\infty$ as $\alpha\to0$. Then
\begin{equation*}
  Z^\alpha(t)
  =\frac{u_{q_\alpha+\lfloor t/\alpha\rfloor}-u^{\dagger}}
  {\sqrt{\alpha}}
\end{equation*}
converges weakly on each bounded time interval to the stationary Ornstein--Uhlenbeck (OU) process
\begin{equation}
  dZ(t)=-aZ(t)\,dt+
  \frac{\varsigma(u^{\dagger})}{V^{\star}}\,dW(t),
  \label{eq:ou}
\end{equation}
where $W$ is a standard Wiener process. In particular,
\begin{equation}
  \frac{u_{q_\alpha}-u^{\dagger}}{\sqrt{\alpha}}
  \ \Rightarrow\
  \mathcal{N}\!\left(
  0,\frac{\varsigma^2(u^{\dagger})}{2a(V^{\star})^2}
  \right).
  \label{eq:small_step_normal}
\end{equation}
\end{proposition}
\begin{IEEEproof}
Let $h(u)=(F(u)-V^{\star})/V^{\star}$ and $e_n=u_n-u^{\dagger}$. The update is
\begin{equation*}
  u_{n+1}=\Pi_{\mathcal{U}}\left(
  u_n+\alpha h(u_n)+\frac{\alpha}{V^{\star}}\varepsilon_n
  \right).
\end{equation*}
We apply the constant-step diffusion theorem in~\cite[Thm.~10.1.1]{kushner2003stochastic} after shifting the recursion beyond its initial transient.

By (A1) and compactness of $\mathcal{U}$, $a_{\min}=\min_{u\in\mathcal{U}}|F'(u)|/V^{\star}>0$. Since $h(u^{\dagger})=0$, the mean value theorem gives $e_nh(u_n)\le-a_{\min}e_n^2$. Projection does not increase the distance to $u^{\dagger}$, and the conditional mean of $\varepsilon_n$ is zero. Consequently,
\begin{equation*}
\begin{aligned}
  \mathbb{E}[e_{n+1}^2]
  &\le (1-2\alpha a_{\min})\mathbb{E}[e_n^2]
  +\alpha^2\Gamma,\\
  \Gamma
  &=\max_{u\in\mathcal{U}}h(u)^2
  +\frac{\omega^2}{(V^{\star})^2}.
\end{aligned}
\end{equation*}
For sufficiently small $\alpha$, iteration yields
\begin{equation*}
  \mathbb{E}[e_n^2]
  \le e^{-2\alpha a_{\min}n}(u_{\max}-u_{\min})^2
  +\frac{\alpha\Gamma}{2a_{\min}}.
\end{equation*}
Define
\begin{equation*}
  p_\alpha=
  \left\lceil\frac{|\log\alpha|}{2\alpha a_{\min}}\right\rceil,
  \qquad
  \tilde u_n^\alpha=u_{p_\alpha+n}.
\end{equation*}
Then, for all sufficiently small $\alpha$,
\begin{equation*}
  \sup_{n\ge0}
  \frac{\mathbb{E}[(\tilde u_n^\alpha-u^{\dagger})^2]}{\alpha}
  \le (u_{\max}-u_{\min})^2+\frac{\Gamma}{2a_{\min}}.
\end{equation*}
Thus, the normalized errors of the shifted recursion form a tight family, verifying (A10.1.3).

We next verify concentration of the shifted interpolated process, as required by (A10.1.2). Fix a neighborhood of $u^{\dagger}$ whose closure lies inside $\mathcal{U}$, and stop the shifted recursion on leaving this neighborhood. Since $h$ and the noise are bounded, individual proposed increments are $O(\alpha)$, so projection is inactive before exit for sufficiently small $\alpha$. Over any fixed rescaled time interval $[0,H]$, the martingale maximal inequality bounds the expected squared maximum of the accumulated noise by $O(\alpha)$. Together with the preceding initial-error bound, Lipschitz continuity of $h$, and a discrete Gronwall inequality, this gives an $O(\alpha)$ bound on the expected squared maximum of the stopped deviation. The exit probability therefore tends to zero, and
\begin{equation*}
  \sup_{0\le t\le H}
  \left|\tilde u_{\lfloor t/\alpha\rfloor}^\alpha-u^{\dagger}\right|
  \xrightarrow{\mathbb{P}}0.
\end{equation*}
Hence the shifted process converges to the constant interior process $u^{\dagger}$, verifying (A10.1.2). For this shifted recursion, the delay indices in (A10.1.2) and (A10.1.3) can both be taken as zero.

The remaining conditions follow from the drift and noise assumptions. Boundedness of $h$ and the noise gives the uniform integrability in (A10.1.1) and the higher-moment bound in (A10.1.7). The conditional mean drift is $h$, which is continuously differentiable and independent of $n$ and $\alpha$, with $h(u^{\dagger})=0$ and $h'(u^{\dagger})=-a<0$. These properties verify (A10.1.4)--(A10.1.6), with Hurwitz matrix $A=-a$. For the shifted martingale difference $\delta\tilde M_n^\alpha=\varepsilon_{p_\alpha+n}/V^{\star}$,
\begin{equation*}
  \mathbb{E}\!\left[
  (\delta\tilde M_n^\alpha)^2\mid\mathcal{F}_{p_\alpha+n}
  \right]
  =\frac{\varsigma^2(\tilde u_n^\alpha)}{(V^{\star})^2}.
\end{equation*}
The uniform moment bound implies $\tilde u_n^\alpha\to u^{\dagger}$ in probability uniformly over $n\ge0$ as $\alpha\to0$. Continuity of $\varsigma^2$ therefore gives convergence of these conditional variances to $\varsigma^2(u^{\dagger})/(V^{\star})^2$. The same convergence holds with the neighborhood indicator required in (A10.1.7), since the probability of lying outside that neighborhood tends to zero.

Finally, let $\tilde q_\alpha=q_\alpha-p_\alpha$. Since $\alpha p_\alpha=O(|\log\alpha|)$, the assumed condition on $q_\alpha$ gives $\alpha\tilde q_\alpha\to\infty$. Applying~\cite[Thm.~10.1.1]{kushner2003stochastic} to the shifted recursion at indices $\tilde q_\alpha$ yields weak convergence of
\begin{equation*}
  \frac{\tilde u_{\tilde q_\alpha+\lfloor t/\alpha\rfloor}^\alpha-u^{\dagger}}
       {\sqrt{\alpha}}
  =Z^\alpha(t)
\end{equation*}
to the stationary solution of \eqref{eq:ou}. Its stationary distribution is $\mathcal{N}\bigl(0,\varsigma^2(u^{\dagger})/[2a(V^{\star})^2]\bigr)$. Evaluation at $t=0$ gives \eqref{eq:small_step_normal}.
\end{IEEEproof}

Proposition~\ref{prop:fluct} gives exact response-time and variance formulas for the affine model. For small $\alpha a$ and a fixed update period, the response time scales as $1/\alpha$, while the standard deviation of $\log\lambda$ scales as $\sqrt{\alpha}$. With an update period of $P$ slots, the approximate time constant is $\tau=P/(\alpha a)$, so the local sensitivity $a$ determines how a given step size translates into adaptation speed. In this model, the mean approaches the target without alternating signs when $0<\alpha a<1$ and alternates around it when $1<\alpha a<2$.

Proposition~\ref{prop:small_step} connects these formulas to the nonlinear projected update. Under its assumptions, the normalized error process converges weakly to the stationary Ornstein--Uhlenbeck process in \eqref{eq:ou} as $\alpha\to0$ after a sufficiently long initial transient. The limiting distribution gives the variance approximation $\alpha\varsigma^2(u^{\dagger})/[2a(V^{\star})^2]$, matching the leading term of \eqref{eq:var} when $\omega^2=\varsigma^2(u^{\dagger})$. The limiting process has time constant $1/a$ in the rescaled time $t=\alpha n$, corresponding to approximately $1/(\alpha a)$ updates. Thus, the small-step analysis establishes that the response-time and fluctuation scaling derived from the affine model also arises in the nonlinear projected update.

For parameter selection, \eqref{eq:var} provides a local approximation when the response is approximately affine and projection events are negligible. Its predicted standard deviation can be compared with the distance $d=\min\{u^{\dagger}-u_{\min},u_{\max}-u^{\dagger}\}$ to assess the relevance of the multiplier bounds. Near a boundary, projection can substantially affect both fluctuations and average tracking error. The condition $0<\alpha a<2$ characterizes stability of the unprojected affine model; the small-step result does not extend this condition to arbitrary step sizes in the nonlinear update. These results inform the parameter-selection procedure in Section~\ref{subsec:operation}, and Section~\ref{subsec:res_closed} evaluates the approximation at the implemented step size.

\subsection{Finite-Window Service Tracking}
\label{subsec:an_average}

The following result gives an exact identity for the average tracking error of the nonlinear projected update over a finite window. Unlike the preceding response-based analyses, it holds for every measurement sequence without assumptions (A1)--(A2).

\begin{proposition}[Finite-window service tracking]
\label{prop:average}
Consider \eqref{eq:dual_log} with constant step $\alpha>0$, fixed target $V^{\star}>0$, and $u_0\in\mathcal{U}$. Define the projection correction at update $n$ as
\[
  z_n=u_{n+1}-u_n-\frac{\alpha}{V^{\star}}(\hat V_n-V^{\star}).
\]
For every $n_0\ge0$ and $N\ge1$,
\begin{equation}
  \begin{aligned}
  &\frac{1}{N}\sum_{n=n_0}^{n_0+N-1}\hat V_n-V^{\star}=\frac{V^{\star}}{\alpha N}\left(
    u_{n_0+N}-u_{n_0}
    -\sum_{n=n_0}^{n_0+N-1}z_n
  \right).
  \end{aligned}
  \label{eq:average}
\end{equation}
The endpoint difference satisfies $|u_{n_0+N}-u_{n_0}|\le u_{\max}-u_{\min}$. Upper-bound clipping gives $z_n<0$, lower-bound clipping gives $z_n>0$, and $z_n=0$ when the proposed update lies in $\mathcal{U}$. If projection is inactive throughout the window, then
\begin{equation}
  \left|
  \frac{1}{N}\sum_{n=n_0}^{n_0+N-1}\hat V_n-V^{\star}
  \right|
  \le\frac{V^{\star}(u_{\max}-u_{\min})}{\alpha N}.
  \label{eq:average_bound}
\end{equation}
\end{proposition}

\begin{IEEEproof}
By the definition of $z_n$,
\[
  u_{n+1}-u_n
  =\frac{\alpha}{V^{\star}}(\hat V_n-V^{\star})+z_n.
\]
Summing over the window and rearranging gives \eqref{eq:average}. Both endpoints lie in $\mathcal{U}$, so their difference is bounded in magnitude by $u_{\max}-u_{\min}$. Projection reduces a proposed update above $u_{\max}$, giving $z_n<0$, and increases one below $u_{\min}$, giving $z_n>0$. Otherwise, $z_n=0$. With no projection corrections, bounding the endpoint difference yields \eqref{eq:average_bound}.
\end{IEEEproof}

Equation~\eqref{eq:average} separates the average tracking error into an endpoint contribution and a projection contribution. The endpoint contribution is $O(1/N)$ because the multiplier range is bounded. Consequently, over growing windows without projection, the average measured degradation approaches the target even if individual measurements and multipliers continue to fluctuate. The same conclusion holds whenever $N^{-1}\sum_{n=n_0}^{n_0+N-1}z_n\to0$. Upper-bound corrections contribute positively to the average tracking error, whereas lower-bound corrections contribute negatively, explaining how range saturation can sustain a deviation from the target.

For update periods of equal duration, the average of $\hat V_n$ equals the slot-average degradation over the same window. With $V^{\star}=V_{\max}-m$ and no projection, \eqref{eq:average_bound} gives the sufficient condition $m\ge V^{\star}(u_{\max}-u_{\min})/(\alpha N)$ for this measured average to satisfy the service requirement. This worst-case condition uses the full operational range, whereas \eqref{eq:average} gives the exact error from the realized endpoint change and projection corrections. Under (A2), the martingale-difference measurement errors have uniformly bounded second moments, so their averages converge to zero almost surely. Thus, for any fixed starting index $n_0$, the averages of $\hat V_n$ and $F(u_n)$ have the same asymptotic behavior.

\subsection{Switching under the Dwell Constraint}
\label{subsec:an_dwell}

The following result bounds cell switching independently of the multiplier trajectory. It requires only the enforced dwell constraint and does not rely on assumptions (A1)--(A2).

\begin{proposition}[Dwell-time switching bound]
\label{prop:dwell}
Suppose successive state transitions of each cell are separated by at least $T_{\mathrm{dwell}}$ slots. In any interval of $L\ge1$ consecutive slots, each cell undergoes at most
\begin{equation}
  N_{\mathrm{tr}}(L)
  =\left\lceil\frac{L}{T_{\mathrm{dwell}}}\right\rceil
  \label{eq:transition_bound}
\end{equation}
state transitions and at most
\begin{equation}
  N_{\mathrm{off}}(L)
  =\left\lceil\frac{N_{\mathrm{tr}}(L)}{2}\right\rceil
  \label{eq:deactivation_bound}
\end{equation}
deactivations, regardless of the multiplier sequence.
\end{proposition}

\begin{IEEEproof}
Partition the interval into $\lceil L/T_{\mathrm{dwell}}\rceil$ consecutive blocks, each containing at most $T_{\mathrm{dwell}}$ slots. Two transitions within the same block would be separated by fewer than $T_{\mathrm{dwell}}$ slots, contradicting the dwell constraint. Each block therefore contains at most one transition, giving \eqref{eq:transition_bound}. Since transitions alternate between deactivation and reactivation, at most $\lceil N_{\mathrm{tr}}(L)/2\rceil$ can be deactivations.
\end{IEEEproof}

Proposition~\ref{prop:dwell} gives a deterministic switching bound that remains valid under online multiplier adaptation. Over a cycle of $T$ slots, the cluster undergoes at most $B N_{\mathrm{tr}}(T)$ transitions and $B N_{\mathrm{off}}(T)$ deactivations. The dwell constraint limits switching frequency but still permits reversals after the dwell time expires. Whether multiplier adaptation increases switching within these bounds is evaluated in Section~\ref{subsec:res_closed} by comparing fixed-multiplier and online operation at matched degradation.

\section{Performance Evaluation}
\label{sec:evaluation}

We evaluate ATTUNE in terms of service-target tracking and energy savings under different operating conditions. Baseline comparisons and ablation studies assess the effectiveness of its training and deployment mechanisms.

\subsection{Setup}
\label{subsec:setup}

\noindent\textit{Scenario:} We evaluate a cluster of $B=7$ small cells serving $100$ UEs, with $70\%$ initially located in two hotspots at non-adjacent outer cells. The hotspot demands peak half a cycle apart. A cycle contains $T=3600$ slots of duration $T_0=1$~s and represents a compressed diurnal profile. ES decisions occur every $K=60$ slots, with a minimum dwell time of $T_{\mathrm{dwell}}=300$ slots. Service degradation is the fraction of UEs in outage or receiving less than $\beta=0.8$ of their demand. We consider $V_{\max}\in\{0.03,0.05,0.10\}$ and adopt the nominal traffic load and remaining network parameters of~\cite{jeong2026accord}.

\noindent\textit{Training:} We train for $1000$ iterations using $16$ parallel environments and five training seeds. The discount factors are $\gamma=0.99$ and $\gamma_{\mathrm{E}}=0.98$, and the generalized advantage estimation parameter is $\lambda_{\mathrm{GAE}}=0.95$, following~\cite{jeong2026accord}. The effective discounted cost horizon is $1/(1-\gamma)=100$ slots, whereas the table and operational updates use cycle-average degradation. For ATTUNE, targets are sampled uniformly from $[V_{\mathrm{lo}},V_{\mathrm{hi}}]=[0.025,0.13]$. The target grid is $\{0.025,0.04,0.06,0.08,0.10,0.13\}$, with table-update parameters $\eta=0.05$, $\delta=0.05$, and $\kappa=0.1$, multiplier bounds $[\underline\lambda,\bar\lambda]=[0.05,200]$, and log-space perturbation standard deviation $\sigma=0.5$. During the first $200$ iterations, multipliers are drawn log-uniformly from $[0.4,15]$. Measurements from the final $50$ warm-up iterations are used for the least-squares initialization described in Section~\ref{subsec:training}.

\begin{table*}[t]
\centering
\caption{Energy-saving difference from specialized ACCORD at matched degradation [percentage points], with 95\% confidence intervals and the number of seeds covering the reference degradation. The final row reports specialized ACCORD energy savings [\%].}
\label{tab:distribution}
\footnotesize
\setlength{\tabcolsep}{4pt}
\begin{tabular}{@{}lcccccc@{}}
\toprule
 & \multicolumn{2}{c}{$V_{\max}=0.03$ ($\bar V=0.031$)} & \multicolumn{2}{c}{$V_{\max}=0.05$ ($\bar V=0.045$)} & \multicolumn{2}{c}{$V_{\max}=0.10$ ($\bar V=0.101$)} \\
\cmidrule(lr){2-3}\cmidrule(lr){4-5}\cmidrule(l){6-7}
Training distribution & Difference & Seeds & Difference & Seeds & Difference & Seeds \\
\midrule
Fixed range $[0.4,15]$   & $+0.92$ $[-0.50,+2.34]$ & 2/5 & $+0.53$ $[0.00,+1.06]$ & 4/5 & $-0.24$ $[-1.69,+0.91]$ & 1/5 \\
Widened range $[0.1,50]$ & $-6.77$ $[-7.86,-5.81]$ & 4/5 & $-8.46$ $[-9.10,-7.59]$ & 4/5 & $-9.21$ $[-10.94,-7.52]$ & 5/5 \\
Adaptive range           & $-0.36$ $[-1.56,+0.80]$ & 5/5 & $-1.13$ $[-1.94,-0.38]$ & 5/5 & $-0.44$ $[-2.06,+0.97]$ & 5/5 \\
Target-uniform (ATTUNE)  & $+0.07$ $[-1.27,+1.45]$ & 5/5 & $-0.32$ $[-1.11,+0.52]$ & 5/5 & $+0.23$ $[-1.30,+1.59]$ & 5/5 \\
\midrule
Specialized ACCORD       & 21.1 $[20.0,22.1]$ & 5/5 & 25.4 $[24.9,25.8]$ & 5/5 & 32.0 $[30.8,33.3]$ & 5/5 \\
\bottomrule
\end{tabular}

\smallskip
\begin{minipage}{\textwidth}
\footnotesize
Reference degradation levels are the five-seed means of the specialized policies; differences include only seeds whose response curves cover the corresponding reference, as counted in the Seeds columns.
\end{minipage}
\end{table*}

\noindent\textit{Evaluation:} Fixed-multiplier response curves are evaluated over $150$ cycles using the protocol of~\cite{jeong2026accord}. Online evaluation uses five parallel trajectories of $30$ cycles each. Network states are reinitialized at the start of each cycle, while multipliers are retained between cycles. Unless otherwise specified, reported averages use the final $15$ cycles of each trajectory. The default operational settings are an update period of one cycle, $\alpha=0.5$, and $\lambda_0=\mathrm{clip}_{\Lambda}(1)$. The step size was selected beforehand using the fixed-range policy. We use $m=0$, so that $V^{\star}=V_{\max}$, unless a target margin is explicitly specified. Open- versus closed-loop comparisons and cross-seed pairing experiments instead initialize the multiplier at the table value corresponding to $V^{\star}$. The observed initial transient in these fixed-condition evaluations lasts $2$--$9$ cycles. For pairs trained under different loads, ATTUNE is additionally trained jointly at $U=75$ with five seeds, using the same demand-input normalization as at $U=100$.

\noindent\textit{Policies compared:} We construct baselines that incorporate the state-augmentation and threshold-conditioning principles summarized in Table~\ref{tab:related} into a common ES--MLB setting. The methods share the base network architecture, PPO training procedure, MLB agent design, and common hyperparameters of~\cite{jeong2026accord}. Differences in policy inputs, training distributions, and deployment updates are specified below.
\begin{itemize}
\item \emph{Specialized}: ACCORD policies trained separately for each service requirement $V_{\max}$~\cite{jeong2026accord}, providing a reference for the energy savings achieved through target-specific training.
\item \emph{Fixed range}: multiplier-conditioned policies based on the state-augmentation framework of~\cite{calvofullana2024state,naderializadeh2022state}, using fixed-distribution multiplier sampling as in~\cite{naderializadeh2022state}. Multipliers are drawn log-uniformly from $[0.4,15]$ or, for the widened variant, $[0.1,50]$, and held fixed within each episode. These baselines assess fixed multiplier distributions within the common ES--MLB training setup.
\item \emph{Adaptive range}: multiplier-conditioned policies whose log-uniform sampling range is adjusted during training so that its endpoints correspond to prescribed service levels. This baseline compares range adaptation with ATTUNE's target-uniform sampling.
\item \emph{ATTUNE}: the target-uniform training procedure of Section~\ref{subsec:training} with the online multiplier update of Section~\ref{subsec:operation}.
\item \emph{Threshold-conditioned}: a baseline following the threshold-conditioning principle of CCPO~\cite{yao2023ccpo}. The ES policy receives a service threshold instead of $\lambda$, while retaining ATTUNE's base architecture, target distribution, and training multiplier table. We evaluate both a fixed input $V^{\star}$ and a feedback-corrected input initialized at $V^{\star}$ and updated once per cycle as $V^{\mathrm{in}}_{n+1}=\mathrm{clip}_{[V_{\mathrm{lo}},V_{\mathrm{hi}}]}\bigl(V^{\mathrm{in}}_n-k_v(\hat V_n-V^{\star})\bigr)$. These variants distinguish the effect of the conditioning variable from that of deployment feedback. We use $k_v=0.156$, selected for a time constant of approximately $8$ cycles based on the fixed-range policy; ATTUNE's multiplier update has a time constant of $5$--$6$ cycles.
\end{itemize}
For separately trained application pairs, we also compare ATTUNE with ACCORD's recalibration and validation procedure~\cite{jeong2026accord}. Under overload, we include ACCORD's traffic-based guard as a reference.

\noindent\textit{Metrics:} Energy saving is measured relative to the all-active configuration, and $\bar V$ denotes average degradation over the reported cycles. A pair is classified as tracking its target if $|\bar V-V^{\star}|\le0.005$ and as service-feasible if $\bar V\le V_{\max}$. Switching is measured by the numbers of deactivations and reversals per cycle, where a reversal is a deactivation followed by reactivation of the same cell within $T_{\mathrm{dwell}}+K$ slots~\cite{jeong2026accord}. Energy savings at matched degradation are obtained by interpolating each policy's fixed-multiplier trade-off curve at the nominal load.

\subsection{Service-Target Coverage}
\label{subsec:res_single}
\label{subsec:res_distribution}

We evaluate whether a single ATTUNE policy can cover multiple service levels with energy savings comparable to target-specific policies, and how the training distribution affects coverage and efficiency. Comparisons with specialized ACCORD use its five-seed mean degradation as the reference, with energy savings obtained by interpolating each augmented policy's fixed-multiplier response curve.

ATTUNE covers all three reference service levels for every training seed while achieving energy savings close to specialized ACCORD (Table~\ref{tab:distribution}). The differences are $+0.07$, $-0.32$, and $+0.23$ percentage points for the references associated with $V_{\max}=0.03$, $0.05$, and $0.10$, respectively, with all confidence intervals containing zero. Thus, each trained ATTUNE policy supports the three evaluated service levels without separate training for each requirement.

\begin{figure}[t]
\centering
\includegraphics[width=\columnwidth]{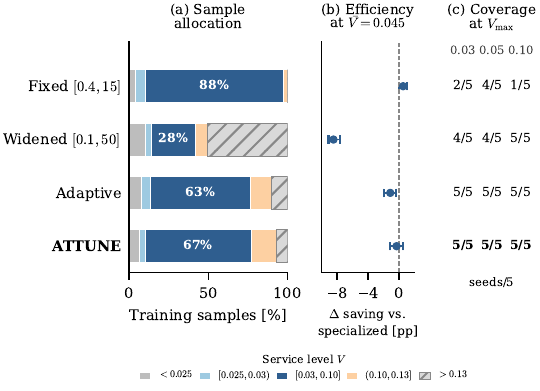}
\caption{Effect of the multiplier training distribution. (a)~Share of training samples grouped by the mean degradation attained by the final policy at the sampled multiplier; the operational service range $[0.03,0.10]$ is highlighted. (b)~Energy-saving difference from specialized ACCORD at $\bar V=0.045$, with 95\% confidence intervals. (c)~Number of seeds whose response curves cover the specialized-policy reference degradation associated with each $V_{\max}$ in Table~\ref{tab:distribution}.}
\label{fig:distribution}
\end{figure}

A fixed multiplier range presents a trade-off between service-level coverage and energy efficiency. The narrow range $[0.4,15]$ concentrates $88\%$ of samples within the operational service range $[0.03,0.10]$ and achieves energy savings close to specialized ACCORD among seeds covering each reference. However, only four of five seeds cover $\bar V=0.045$, and only one or two cover the other reference levels (Fig.~\ref{fig:distribution}(c)). Widening the range to $[0.1,50]$ improves coverage but allocates $51\%$ of samples above $V=0.13$ and only $28\%$ within the operational service range. Its energy savings are $6.8$--$9.2$ percentage points below specialized ACCORD across the three references, including a difference of $-8.46$ percentage points at $\bar V=0.045$ (Fig.~\ref{fig:distribution}(b)). Thus, widening the multiplier range alone does not preserve energy efficiency.

Adaptive-range and target-uniform sampling both cover all three reference levels for every seed. ATTUNE's mean energy savings are closer to those of specialized ACCORD at all three references. At $\bar V=0.045$, adaptive-range sampling gives a difference of $-1.13$ percentage points with confidence interval $[-1.94,-0.38]$, whereas ATTUNE gives $-0.32$ with interval $[-1.11,+0.52]$. These intervals compare each method with specialized ACCORD; they do not establish a statistically significant difference between ATTUNE and adaptive-range sampling.

The distinction between the two adaptive methods lies in sample allocation within the multiplier range. Adaptive-range sampling adjusts the endpoints but retains a log-uniform distribution, whose induced service-level density depends on the response slope as characterized by Proposition~\ref{prop:density}. ATTUNE instead samples service targets uniformly and learns their multiplier mapping. Fig.~\ref{fig:distribution}(a), which evaluates sampled multipliers using the final policies, shows that $66.6\%$ of ATTUNE samples fall within $[0.03,0.10]$, close to the $66.7\%$ target-sampling probability implied by $[0.025,0.13]$; adaptive-range sampling allocates approximately $63\%$ to this interval. The achieved distribution remains subject to table approximation, perturbation, and clipping. Beyond sampling, ATTUNE's learned table provides target-specific initial multipliers and response-sensitivity estimates for the operational parameter selection in Section~\ref{subsec:operation}.

The training distribution also affects the MLB partner, even though MLB does not observe the multiplier. With the narrow-range ES actor held fixed, replacing its MLB partner with the widened-range MLB actor reduces energy saving at $\bar V=0.045$ by $3.0$ percentage points, with confidence interval $[-4.35,-1.88]$. The target-uniform MLB actor yields a smaller difference of $-0.57$ percentage points, with confidence interval $[-1.46,+0.29]$. These results support training the pair on activation patterns relevant to the intended service range.

\subsection{Closed-Loop Tracking and Switching}
\label{subsec:res_closed}
\label{subsec:res_limits}

The online update tracks all three service targets across the five training seeds. Starting from $\lambda_0=\mathrm{clip}_{\Lambda}(1)$, every seed achieves average degradation within $\pm0.005$ of each target. The means across seeds are $0.030$, $0.049$, and $0.098$ for targets of $0.03$, $0.05$, and $0.10$, respectively.

The remaining average tracking error is accounted for by the endpoint and projection terms in Proposition~\ref{prop:average}. In windows without projection, the endpoint change over the final $14$ updates is approximately $10\%$ of the log-multiplier range, accounting for trajectory-average deviations of up to approximately $10\%$ of $V^{\star}$. The endpoint contribution is bounded by a term that decreases as $1/N$, and numerical residuals of the identity are below $10^{-16}$ in every run. Separate load-variation runs initialized at $\lambda_0=\mathrm{clip}_\Lambda(1)$, rather than at the table value used in Fig.~\ref{fig:load}, illustrate the projection contribution: at $U=50$, lower-bound saturation occurs in $24$ of $25$ trajectories and degradation remains below the target; at $U=125$ and $150$, upper-bound saturation accompanies degradation above the target. The accumulated projection corrections account for these sustained deviations.

The measured multiplier fluctuations support the step-size trade-off characterized by Proposition~\ref{prop:fluct}. For the fixed-range policy trained on $[0.4,15]$, an eightfold step-size increase produces an approximately $3.5$-fold increase in log-multiplier fluctuations, compared with $3.2$ predicted by \eqref{eq:var}, while the local approximation overestimates their magnitude by 15--30\%.

For ATTUNE at $V^{\star}=0.05$, adjacent entries of the trained table give a median sensitivity of $a=0.33$ across seeds, approximately consistent with the fixed-multiplier response curves. With $\alpha=0.5$ and a one-cycle update period, the table-based estimate gives a time constant of approximately $6$ cycles; the response curves give approximately $5$ cycles and a median $\alpha a\approx0.2$. The predicted stationary standard deviation of $\log\lambda$ is $0.52$; over $15$- and $30$-cycle windows, the predictions are $0.39$ and $0.45$, against measured values of $0.41$ and $0.46$. These comparisons assess the local approximation at the implemented step size, while Proposition~\ref{prop:small_step} establishes its leading-order connection to the nonlinear update as $\alpha\to0$. The nearer operational boundary lies $4.6$--$8.9$ predicted standard deviations from the target operating point at $V^{\star}=0.05$, compared with a median of $2.5$ at $V^{\star}=0.03$, highlighting the relevance of boundary proximity at stringent targets.

\begin{table*}[t]
\centering
\caption{Cell switching: deactivations and reversals per cycle, and reversal ratios.}
\label{tab:switching}
\footnotesize
\setlength{\tabcolsep}{4pt}
\begin{tabular}{@{}l ccc ccc ccc@{}}
\toprule
 & \multicolumn{3}{c}{$V_{\max}=0.03$ ($\bar V=0.031$)} & \multicolumn{3}{c}{$V_{\max}=0.05$ ($\bar V=0.045$)} & \multicolumn{3}{c}{$V_{\max}=0.10$ ($\bar V=0.101$)} \\
\cmidrule(lr){2-4}\cmidrule(lr){5-7}\cmidrule(l){8-10}
Method & Off & Rev. & Share [\%] & Off & Rev. & Share [\%] & Off & Rev. & Share [\%] \\
\midrule
Rule-based pair~\cite{jeong2026accord} & 6.66 & 4.47 & 67 [52, 72] & 11.41 & 9.08 & 80 [76, 82] & 17.68 & 15.78 & 89 [88, 90] \\
Specialized ACCORD & 7.88 & 2.82 & 36 [33, 39] & 8.42 & 2.71 & 32 [28, 36] & 9.02 & 2.63 & 29 [28, 31] \\
ATTUNE, fixed $\lambda$ & 9.46 & 3.83 & 41 [38, 44] & 8.86 & 3.04 & 34 [32, 37] & 6.84 & 1.56 & 23 [20, 25] \\
ATTUNE, online update & 9.30 & 3.84 & 41 [39, 44] & 8.71 & 2.90 & 33 [30, 36] & 7.02 & 1.75 & 25 [21, 29] \\
\bottomrule
\end{tabular}

\smallskip
\begin{minipage}{\textwidth}
\footnotesize
Off: deactivations per cycle; Rev.: deactivations followed by reactivation within $T_{\mathrm{dwell}}+K$ slots; Share: reversal-to-deactivation ratio, with 95\% confidence intervals. The $\bar V$ in each header is the specialized-policy reference. The rule-based pair and fixed-multiplier ATTUNE are interpolated at this degradation along the threshold and multiplier, respectively. Online ATTUNE uses $V^{\star}=V_{\max}$ and attains $\bar V=0.030$, $0.049$, and $0.098$ over the final 15 of 30 cycles.
\end{minipage}
\end{table*}

Online multiplier adaptation produces no statistically significant increase in switching in the matched-degradation comparisons. Table~\ref{tab:switching} reports switching near the specialized-policy reference levels: the rule-based pair reverses $67\%$--$89\%$ of deactivations, compared with $41\%$, $33\%$, and $25\%$ for online ATTUNE. In a separate seed-wise comparison at the degradation attained online, rather than at the reference levels of Table~\ref{tab:switching}, differences between online and fixed-multiplier ATTUNE in deactivations and reversals range from $-0.19$ to $+0.15$ per cycle, with confidence intervals containing zero. Proposition~\ref{prop:dwell} gives a deterministic upper bound on switching, while these comparisons assess the additional switching associated with multiplier adaptation.

The higher reversal count at the strictest target is observed with both fixed and updated multipliers. At $V_{\max}=0.03$, ATTUNE reverses approximately one additional deactivation per cycle relative to the specialized policy, despite comparable energy savings. Handover rates remain close: $0.0123$ $[0.0117,0.0129]$ for ATTUNE and $0.0126$ $[0.0122,0.0130]$ for the specialized policy. Similar reversal counts with fixed and updated multipliers suggest that the additional reversals are associated with the learned policy rather than multiplier adaptation. 

\subsection{Adaptation across Operating Conditions}
\label{subsec:res_openclosed}
\label{subsec:res_deploy}

We evaluate whether deployment feedback maintains target tracking when traffic load or the MLB partner differs from the training conditions, and compare multiplier and threshold conditioning. All ATTUNE network parameters remain fixed. Table~\ref{tab:openclosed} compares fixed inputs with feedback-adjusted inputs. Open-loop operation uses the table multiplier $\lambda(V^{\star})$ or the prescribed threshold $V^{\star}$, while closed-loop operation adjusts the corresponding input using measured degradation.

Measurement feedback improves target tracking under both nominal and changed operating conditions. For the original policy pairs, fixed multiplier and threshold inputs satisfy the tracking tolerance in $6/15$ and $8/15$ cases, respectively, compared with $15/15$ for ATTUNE. Under load variation over $U\in\{50,75,100\}$, both open-loop variants satisfy the tolerance in $3/15$ cases, compared with $9/15$ for ATTUNE. At $U=50$ and $75$, open-loop degradation remains between $0.012$ and $0.016$, substantially below the target of $0.05$. The energy savings in Table~\ref{tab:openclosed} are measured at the attained degradation and therefore do not represent matched-degradation efficiency comparisons.

\begin{table*}[t]
\centering
\caption{Open- and closed-loop performance: target tracking, service feasibility, deviations exceeding $0.005$, and mean energy saving.}
\label{tab:openclosed}
\footnotesize
\setlength{\tabcolsep}{4pt}
\begin{tabular}{@{}llcccc@{}}
\toprule
Condition & Metric & \multicolumn{2}{c}{Multiplier input} & \multicolumn{2}{c}{Threshold input} \\
\cmidrule(lr){3-4}\cmidrule(l){5-6}
  & & Closed (ATTUNE) & Open & Open (CCPO-style) & Closed \\
\midrule
Same pairs & Within $\pm0.005$ & 15/15 & 6/15 & 8/15 & 14/15 \\
($V^{\star}\in\{0.03,0.05,0.10\}$) & $\bar V\le V_{\max}$ & 14/15 & 11/15 & 9/15 & 8/15 \\
 & Above / below & 0/0 & 2/7 & 3/4 & 0/1 \\
 & Saving [\%] & 26.2 & 25.4 & 25.5 & 25.8 \\
\midrule
Load & Within $\pm0.005$ & 9/15 & 3/15 & 3/15 & 7/15 \\
($U\in\{50,75,100\}$) & $\bar V\le V_{\max}$ & 15/15 & 13/15 & 15/15 & 14/15 \\
 & Above / below & 0/6 & 1/11 & 0/12 & 0/8 \\
 & Saving [\%] & 37.9 & 34.3 & 33.6 & 37.3 \\
\midrule
Cross pairs & Within $\pm0.005$ & 20/20 & 6/20 & 11/20 & 20/20 \\
 & $\bar V\le V_{\max}$ & 17/20 & 10/20 & 17/20 & 17/20 \\
 & Above / below & 0/0 & 6/8 & 0/9 & 0/0 \\
 & Saving [\%] & 25.2 & 25.8 & 24.3 & 24.9 \\
\bottomrule
\end{tabular}

\smallskip
\begin{minipage}{\textwidth}
\footnotesize
CCPO-style denotes a threshold-conditioned baseline adapted to the common ES--MLB training setup~\cite{yao2023ccpo}. Multiplier inputs are initialized from each ES actor's table at $V^{\star}$; threshold inputs start at $V^{\star}$. Above / below counts cases with $\bar V>V^{\star}+0.005$ or $\bar V<V^{\star}-0.005$, respectively. Savings are averaged at the attained degradation, without matching $\bar V$ across methods.
\end{minipage}
\end{table*}

\begin{figure}[t]
\centering
\includegraphics[width=\columnwidth]{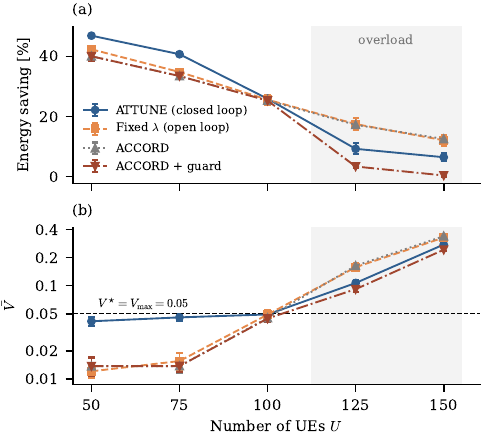}
\caption{Energy saving (a) and degradation (b) under load variation, with $V^{\star}=V_{\max}=0.05$; shaded loads indicate overload. ATTUNE and the fixed-multiplier baseline start from $\lambda(0.05)$ and report the final 15 of 30 cycles. ACCORD and its traffic-based guard are from~\cite{jeong2026accord}. Error bars show 95\% confidence intervals.}
\label{fig:load}
\end{figure}

Under lighter traffic, ATTUNE uses the available service margin to increase energy savings. Fig.~\ref{fig:load} evaluates policies trained at $U=100$ under different UE counts, with per-UE demand unchanged. At $U=50$ and $75$, ATTUNE saves $46.9\%$ and $40.7\%$ at degradation levels of $0.041$ and $0.046$, respectively. Corresponding savings are $42.4\%$ and $34.8\%$ with a fixed multiplier, and $40.1\%$ and $33.5\%$ with ACCORD; both approaches maintain degradation below $0.016$. At $U=50$, lower-bound saturation limits further adjustment toward the target.

\begin{figure}[t]
\centering
\includegraphics[width=\columnwidth]{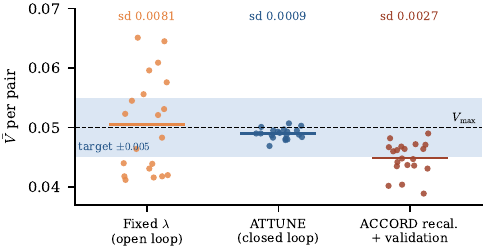}
\caption{Average degradation of $20$ cross-seed application pairs with $V^{\star}=V_{\max}=0.05$. The shaded band indicates the tracking tolerance of $\pm0.005$. ACCORD uses pair-specific recalibration and validation~\cite{jeong2026accord}.}
\label{fig:cross}
\end{figure}

\begin{table}[t]
\centering
\caption{Pairs trained under different loads, operated at $U=100$ with $V^\star=V_{\max}=0.05$ (20 cross-seed pairs each).}
\label{tab:loadpairs}
\footnotesize
\setlength{\tabcolsep}{3pt}
\begin{tabular}{@{}lcccccc@{}}
\toprule
 & \multicolumn{4}{c}{Closed loop (ATTUNE)} & \multicolumn{2}{c}{Open loop} \\
\cmidrule(lr){2-5}\cmidrule(l){6-7}
Pair (training load) & $\pm0.005$ & $\bar V$ & Saving & Rev.\ share & $\pm0.005$ & $\bar V$ \\
\midrule
Same load (100/100) & 20/20 & 0.049 & 25.2\% & 34\% & 6/20 & 0.051 \\
ES 100, MLB 75 & 18/20 & 0.052 & 22.6\% & 40\% & 0/20 & 0.077 \\
ES 75, MLB 100 & 12/20 & 0.055 & 22.4\% & 51\% & 0/20 & 0.135 \\
\bottomrule
\end{tabular}
\end{table}

Under overload, multiplier adaptation reduces degradation but does not restore the service target. All evaluated methods, including the all-active configuration, exceed the requirement. ATTUNE attains degradation of $0.107$ and $0.276$ at $U=125$ and $150$, compared with $0.158$ and $0.328$ for a fixed multiplier, with correspondingly lower energy savings. ACCORD's traffic-based guard attains degradation of $0.092$ and $0.243$ by suspending deactivation. The guard provides a separate mitigation mechanism, while persistent upper-bound saturation indicates that further multiplier increases within $\Lambda$ are unavailable to correct the remaining degradation excess.

Feedback also maintains target tracking when separately trained ES and MLB agents are paired. The fixed multiplier violates the service requirement in $10/20$ cross-seed pairs, including six with degradation above $V^{\star}+0.005$. ATTUNE keeps all $20$ pairs within the tracking tolerance without retraining (Fig.~\ref{fig:cross}). The across-pair standard deviation of $\bar V$ is $0.0009$, compared with $0.0027$ after ACCORD recalibration and $0.0081$ with fixed multipliers. Three ATTUNE pairs exceed the requirement by at most $0.0007$ when $m=0$; setting $V^{\star}=0.048$ yields service-feasible operation for all $20$ pairs. These results distinguish target tracking from strict service feasibility and demonstrate the role of the target margin. Recalibrated ACCORD achieves $26.0\%$ energy saving at average degradation $0.045$, whereas ATTUNE achieves $25.2\%$ with target $0.05$. Partner-specific recalibration therefore achieves higher savings and lower degradation in this comparison, while ATTUNE adapts without updating either actor's network parameters.

Measurement feedback improves tracking with either conditioning variable, but the resulting policies differ in their service response and energy efficiency. Feedback-corrected threshold conditioning meets the tracking tolerance in $14/15$ original-pair cases and all $20$ cross-seed cases (Table~\ref{tab:openclosed}). Under overload, however, it reaches its most conservative input and attains degradation of $0.126$ and $0.292$ at $U=125$ and $150$, compared with $0.107$ and $0.276$ for ATTUNE. At matched degradation, it also achieves approximately $0.5$ percentage points lower energy saving. Thus, both conditioning approaches benefit from deployment feedback, while the evaluated multiplier-conditioned policy achieves lower degradation under overload and higher energy savings at matched degradation.

Applications may also be trained under different loads. Pairing applications trained at $U=75$ with those trained at $U=100$ and operating them at $U=100$, the fixed multiplier misses the tolerance in all pairs in both directions, with degradation of $0.077$ and $0.135$ (Table~\ref{tab:loadpairs}). The online update brings $18$ and $12$ of $20$ pairs within the tolerance, at about $2.7$ percentage points lower saving than same-load pairs and with more reversals. When the ES application is trained under the lighter load, its operating range does not cover the multiplier required at the heavier deployment load for two of five seeds, which saturate at the upper bound as in Proposition~\ref{prop:dual}. These results highlight the importance of covering anticipated deployment loads during ES training. Expanding the operational multiplier range alone does not guarantee target attainment, because the policy response outside the training range has not been established.

\begin{figure}[t]
\centering
\includegraphics[width=\columnwidth]{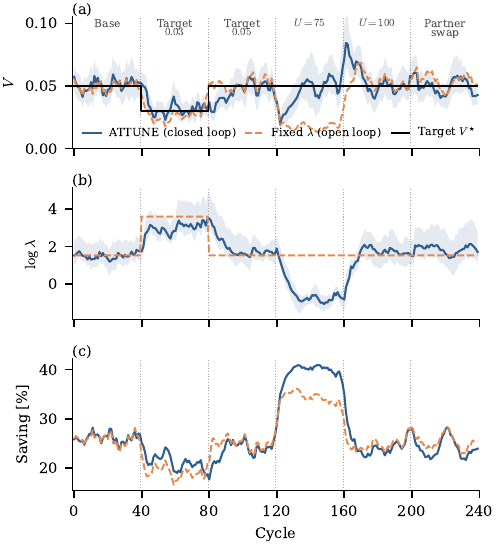}
\caption{Adaptation to sequential changes at 40-cycle intervals: (a)~degradation as a five-cycle moving average and the prescribed target; (b)~log-multiplier; (c)~energy saving. Curves show five-seed means; shading shows the ATTUNE seed range.}
\label{fig:series}
\end{figure}

\subsection{Adaptation to Sequential Changes}
\label{subsec:res_dynamic}

We evaluate whether ATTUNE maintains target tracking through successive changes in the service target, traffic load, and MLB partner. Each training seed is evaluated over a $240$-cycle run comprising six $40$-cycle segments. Following the nominal segment, the target is reduced to $0.03$ and restored to $0.05$, the load is reduced to $U=75$ and restored, and the MLB agent is replaced by a cross-seed partner. The multiplier is retained across segment boundaries, while network states are reinitialized each cycle as in the preceding evaluations. The fixed-multiplier baseline changes its table input only when the prescribed target changes.

ATTUNE adapts to successive changes without resetting the multiplier or retraining the policies. Over the final $10$ cycles of each segment, at least four of five seeds satisfy the tracking tolerance (Fig.~\ref{fig:series}). The fixed-multiplier baseline adjusts its input for prescribed target changes but provides no feedback correction for load or partner changes. After the load reduction, its degradation remains at $0.019$, with no seed within tolerance, whereas ATTUNE achieves approximately $41\%$ energy saving, compared with approximately $35\%$ for the baseline. After partner replacement, all five ATTUNE seeds satisfy the tolerance, compared with one seed for the baseline.

Adaptation takes longer when the new operating point requires a larger multiplier adjustment. Tracking is assessed over $10$-cycle windows containing only post-change measurements, so the earliest reporting time is $10$ cycles after a change. The median time to satisfy the tolerance is $11$ cycles after partner replacement and $18$--$19$ cycles after load changes and target restoration, which require log-multiplier changes of approximately $2$. These longer transients are qualitatively consistent with the finite update increments discussed in Section~\ref{subsec:operation}, although the reported tracking times also include the evaluation window. In an otherwise identical experiment with $20$-cycle segments, only two of five seeds satisfy the tracking tolerance over the final $10$ cycles after target restoration and after load reduction. Both segments therefore fall below the criterion of at least four successful seeds, with substantial transients remaining in the evaluation windows.

These results highlight the need to select the update period and step size according to the time scale of operating-condition changes. If one compressed cycle is interpreted as one physical day, the observed $18$--$19$-cycle tracking time corresponds to approximately three weeks. This illustrative conversion depends on the cycle mapping; the operational parameters must therefore be selected for the intended response time and measurement variability.

\section{Conclusion}
\label{sec:conclusion}

We proposed ATTUNE to adapt a trained ES--MLB policy pair to different service targets and operating conditions without retraining. ATTUNE combines target-uniform multiplier sampling with measurement-based feedback that adjusts the ES policy input while keeping both actors' parameters fixed. The analysis characterizes service-target coverage and tracking dynamics, providing guidance for selecting the update step size and period to balance adaptation speed and multiplier fluctuations. Simulations show energy savings comparable to target-specific policies and approximately seven percentage points greater savings than ACCORD under light load. All 20 cross-seed pairs trained under the same load track their targets within $\pm0.005$, and a target margin yields service-feasible operation for all these pairs. These results support multiplier feedback as a practical mechanism for adapting coexisting applications without policy retraining. Future work will improve generalization across applications trained under different traffic conditions, investigate service guarantees, and extend ATTUNE to multiple interacting applications.

\bibliographystyle{IEEEtran}
\bibliography{IEEEabrv,refs_sa}

\end{document}